\documentclass[runningheads]{llncs}
\usepackage{graphicx}
\usepackage{amsmath,amssymb} 
\usepackage{color}
\usepackage{caption}
\usepackage{multirow}

\usepackage{multido}
\usepackage{kotex}
\usepackage{mathtools}

\newcommand{\bluenote}[1]{{\color{blue}#1}}
\newcommand{\rednote}[1]{{\color{red}#1}}
\newcommand{\rom}[1]{\uppercase\expandafter{\romannumeral #1\relax}}

\newcommand{\describeContent}[1]{%
\begingroup%
\let\thefootnote\relax%
\footnotetext{#1}%
\endgroup%
}

\begin{document}
\pagestyle{headings}
\mainmatter

\def\ACCV20SubNumber{761}  

\title{Pose Correction Algorithm for Relative Frames between Keyframes in SLAM}


\titlerunning{Pose Correction Algorithm}
\authorrunning{Y. Jang, H. Shin, and H. Jin Kim}

\author{Youngseok Jang$^{1}$, Hojoon Shin$^{1}$, and H. Jin Kim$^{1*}$}

\institute{${}^{1, 1*}$ Dep. of Mechanical and Aerospace Engineering, Seoul National University\\
duscjs59@gmail.com, asdwer20@gmail.com, hjinkim@snu.ac.kr}

\maketitle

\begin{abstract}
With the dominance of keyframe-based SLAM in the field of robotics, the relative frame poses between keyframes have typically been sacrificed for a faster algorithm to achieve online applications. However, those approaches can become insufficient for applications that may require refined poses of all frames, not just keyframes which are relatively sparse compared to all input frames. This paper proposes a novel algorithm to correct the relative frames between keyframes after the keyframes have been updated by a back-end optimization process. The correction model is derived using conservation of the measurement constraint between landmarks and the robot pose. The proposed algorithm is designed to be easily integrable to existing keyframe-based SLAM systems while exhibiting robust and accurate performance superior to existing interpolation methods. The algorithm also requires low computational resources and hence has a minimal burden on the whole SLAM pipeline. We provide the evaluation of the proposed pose correction algorithm in comparison to existing interpolation methods in various vector spaces, and our method has demonstrated excellent accuracy in both KITTI and EuRoC datasets.
\end{abstract}

\section{Introduction}
\label{introduction}
\describeContent{${}^{1}$ These authors contributed equally to this manuscript.\\${}^{1*}$ Corresponding author}

Simultaneous localization and mapping (SLAM) has been the focus of numerous research in the field of robotics. SLAM involves estimating the ego-motion of a mobile robot while simultaneously reconstructing the surrounding environment. To this end, visual sensors and laser scanners have been commonly used to perceive the surrounding environment. Vision sensors, in particular, have been most widely adopted due to the wealth of visual information that they can provide at a comparatively low price point. Hence, a vast portion of the SLAM research has been conducted with vision sensors such as monocular, stereo cameras, or RGB-D sensors \cite{mur2017orb,engel2017direct,forster2016svo,engel2014lsd}.

Operating back-end refinement systems such as pose graph optimization (PGO) or bundle adjustment (BA) on all frames can become taxing especially in large-scale environments. To reduce computation time while preserving performance, most modern visual SLAM algorithms adopt keyframe-based approaches which refine only keyframes that contain useful information for SLAM. In other words, keyframe-based SLAM approaches effectively filters the input measurements so that only those that contain significant changes are used in the refinement process, resulting in shorter computation time and local minima avoidance. They allow for the robust estimation of poses and reconstruction of the surrounding map in real-time. 


While keyframe-based SLAM methods have dominated SLAM research, they refine the keyframe poses and do not propagate the corrections to the relative frames between keyframes. Because such systems can only make use of selected keyframes that are relatively sparse compared to the raw input measurements, they are not suitable for applications that require corrected poses at high frequency.
In particular, multi-robot systems that utilize inter-robot relative poses to integrate multiple observations from team robots require a high robot pose density that existing keyframe-based SLAM methods cannot provide. Therefore, an algorithm that can correct poses of relative frames each time the keyframe poses are updated by the back-end of keyframe-based SLAM is required.

Some attempts to correct the relative poses between keyframes have been made in past works using hierarchical PGO. Hierarchical PGO involves dividing the full pose graph into subgraphs that contain representative keyframes called keynodes. The back-end refinement process is conducted only on these selected keynodes and propagated down the hierarchy. The propagation is usually done through either optimization methods or non-optimization methods such as interpolation. Optimization-based correction methods \cite{pinies2007scalable,pinies2008large,suger2014approach} exhibit high accuracy but requires long computation time, making them difficult to operate in real-time. Non-optimization methods \cite{grisetti2007efficient,grisetti2010hierarchical,droeschel2018efficient}, on the other hand, either treat each subgraph as a rigid body or convert the 3D pose into a vector space and interpolates within the given space, allowing for extremely fast operation. However, in such methods, the interpolation factor can become numerically sensitive when the change in a given axis is small. Furthermore, because the method does not consider the measurement constraints between poses, the correction can potentially break these constraints. These two methods will be further discussed in Section \ref{related work}.

This paper proposes a pose correction algorithm for relative frames between keyframes. The algorithm requires just the estimated pose output from a SLAM system to operate, meaning that it can be easily integrated into any existing keyframe-based SLAM methods. The generated pose correction also preserves measurement constraints such as image coordinates of visual features without using optimization, enabling fast computation. The proposed algorithm is compared to existing interpolation-based correction methods in various vector spaces and demonstrates superior accuracy and computation time.

The remainder of this paper is structured as follows. The next section reviews related works regarding pose correction and the problem setup and notation are provided in Section 3. Section 4 describes the proposed pose correction algorithm using measurement constraints. The evaluation results using KITTI and EuRoC datasets are presented in Section 5, and the conclusion of this paper is provided in Section 6.

\section{Related Work}
\label{related work}

This section of the paper discusses the existing work regarding hierarchical PGO and the interpolation in various vector spaces. As mentioned above, distribution of the corrections down to the lower levels of the hierarchy in previous attempts have either used optimization methods or non-optimization methods. This paper will henceforth refer to the former as non-naïve methods and the latter as naïve methods.

The non-naïve approaches to hierarchical PGO involve propagating the refinements of the keyframes to the relative frames through optimization methods. \cite{pinies2007scalable,pinies2008large} proposed separating the full pose graph into sequentially generated and conditionally independent subgraphs. Pose corrections can be conducted by propagating the error from the most recent subgraph. \cite{suger2014approach} followed a similar approach but optimized each subgraph independently. While optimizing each hierarchical subgraph is guaranteed to yield accurate results, the procedure requires high computation times and is not suitable for large-scale SLAM applications. Furthermore, because the implementation requires fundamental changes in the SLAM algorithm itself, it is very difficult to integrate such methods into existing keyframe-based SLAM algorithms without affecting the functionality of the algorithms.  

The naïve methods simplify the constraints between relative frames to propagate the corrections. Interpolation is the most common example of such simplification methods. However, due to the nature of interpolations, if the rate of change between poses are small, the interpolation factor can become numerically sensitive, resulting in extreme values. Furthermore, the accuracy of the methods also suffers, as the interpolation is only concerned with the keyframe poses and does not consider the measurement constraints present in the pose graph. There have been past attempts to develop alternative methods to interpolation. \cite{grisetti2007efficient} proposed an algorithm that utilized the quaternion spherical linear interpolation (slerp) algorithm developed in \cite{shoemake1985animating} to distribute the pose correction to each frame along the traveled path. However, the method requires the covariance of the edges between nodes and also assumes spherical covariances to compute the interpolation factor. Computing the covariance accurately is very difficult and further assumption of spherical covariance that is not guaranteed SLAM applications can further exacerbate the error. \cite{grisetti2010hierarchical} proposed a method where the correction was propagated to the subgraph by treating each subgraph as a rigid body. This simplification assumes that each relative frame receives the same correction and also ignores the measurement constraints between relative frames. More recently, \cite{droeschel2018efficient} proposed a LiDAR-based online mapping algorithm that treats individual scans as subgraphs and propagates the corrections between scan poses using B-spline interpolation. However, this method discards the measurement constraints of relative frames between keyframes and does not hold true if the path generated by the robot does not follow a B-spline trajectory.

The interpolation and linearization of various vector spaces that were used in the naïve methods have also been studied extensively \cite{stuelpnagel1964parametrization,shuster1993survey,barfoot2011pose,blanco2010tutorial}. Pose corrections require the frame poses to be expressed in SE(3). While the translation component of the SE(3) matrix can be readily interpolated, interpolating the rotation matrix may break the SO(3) constraint that defines the rotation matrix. Hence, the rotation matrix must be converted to a vector space in the form of Euler angles, quaternions, so(3), or rotation axis and angle. In this paper, the numerical robustness of these manifolds was tested for their application in the interpolation of robot poses.

\begin{figure}[t]
\centering
\includegraphics[width=100mm]{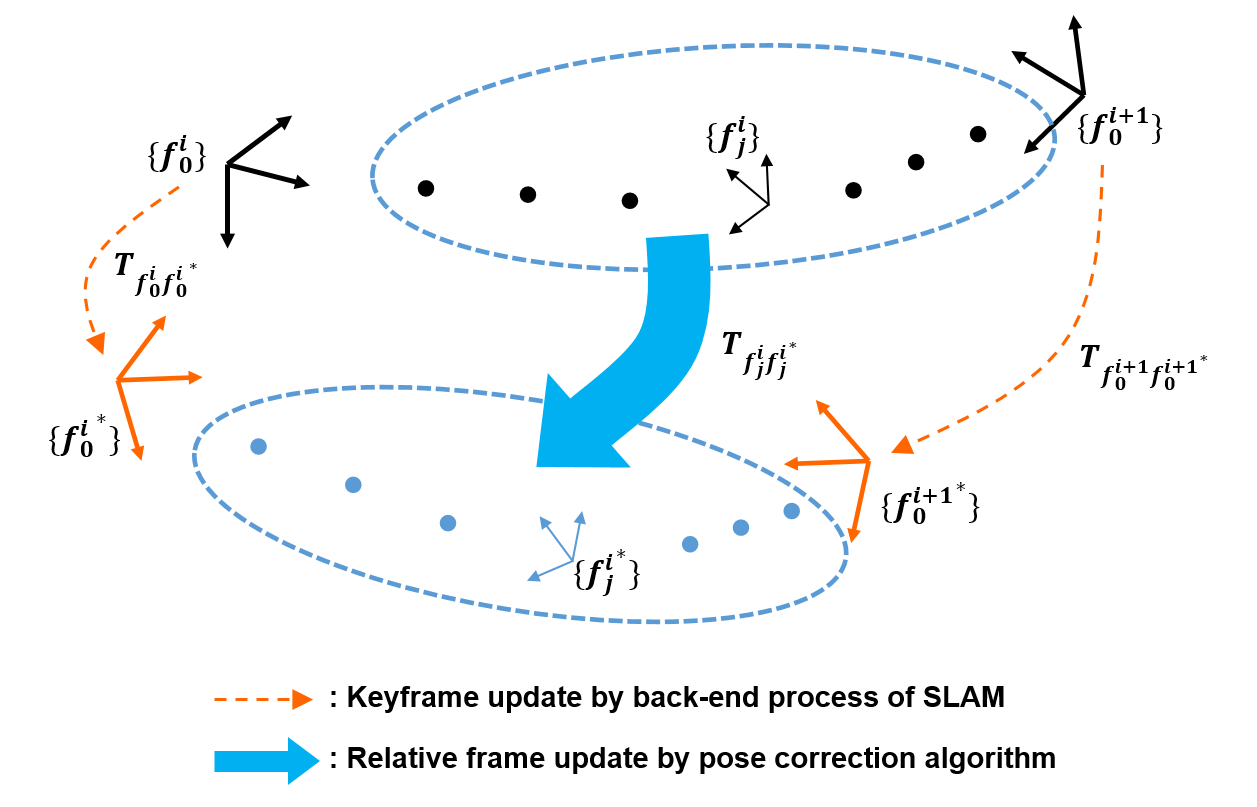} 
\caption{
The refinement process for keyframes and relative frames. The pose correction algorithm for relative frames is triggered each time keyframes are updated by back-end of SLAM.
}
\label{fig:coordinates}
\end{figure}

\section{Problem Statement}
\label{problem statement}
As mentioned previously, most high-performance SLAM algorithms only refine the keyframe poses, which may be too sparse for certain applications. Hence, this paper proposes a fast and easily integrable pose correction algorithm for relative frames between keyframes. 
The previous approaches have typically used interpolation-based correction methods to achieve fast computation for online robot applications. However, such methods are inherently limited by their numerical sensitivity under singular cases involving small changes in a select axis, and may potentially break measurement constraints even under non-singular conditions. The proposed algorithm is not only capable of preserving measurement constraints under most circumstances, but can also robustly correct poses under singular conditions.

Fig. \ref{fig:coordinates} depicts the correction of relative frames between keyframes when the keyframes have been updated by a refinement process such as PGO or BA. $\{f^{i}_{0}\}$ and $\{f^{i}_{j}\}$ are the coordinates of the $i^{\text{th}}$ keyframe and the $j^{\text{th}}$ relative frame connected to the $i^{\text{th}}$ keyframe, respectively. The updated keyframe and the corrected relative frame are denoted as $\{{f^{i}_{0}}^{*}\}$ and $\{{f^{i}_{j}}^{*}\}$. $T_{f_{1}f_{2}}$ is the SE(3) transformation from the $\{f_{1}\}$ coordinate frame to the $\{f_{2}\}$ coordinate frame. The aim is to approximate the relative frame correction transformation $ T_{f_{j}^{i}{f_{j}^{i}}^{*}}$ given the keyframe update transformations $ T_{f_{0}^{i}{f_{0}^{i}}^{*}}$ and $T_{f_{0}^{i+1}{f_{0}^{i+1}}^{*}}$.

Interpolation approaches are typically used to correct the relative poses between keyframes. However, element-wise interpolation of a matrix in the SE(3) may break the rotation matrix SO(3) constraints $\left( \text{det}(R) = +1 \; \text{and} \; R^{\text{T}}R = I \right)$. To prevent this, the SE(3) matrix should first be converted into a vector. The general equation for the interpolation of an SE(3) matrix after the conversion to a vector is as follows:
\begin{align}
    x_{{f_{0}^{i}}^{*}{f_{j}^{i}}^{*}} = x_{f_{0}^{i}f_{j}^{i}} + ( x_{{f_{0}^{i}}^{*}{f_{0}^{i+1}}^{*}} - x_{f_{0}^{i}f_{0}^{i+1}} ) \frac{x_{f_{0}^{i}f_{j}^{i}}}{x_{f_{0}^{i}f_{0}^{i+1}}}
\label{eqn: interp}
\end{align}
where $x_{f_{1}f_{2}} = f \left( T_{f_{1}f_{2}} \right), \; f:\text{SE(3)} \;\; {\mapsto} \;\; {\mathbb{R}}^{n \times 1}$, and $n$ is the dimension of the transformed vector space. The above equation was used to interpolate poses in a variety of vector spaces and the results of the interpolation served as the baseline for comparison with the proposed algorithm. The spaces formed by XYZ and the translation portion of the se(3) were used to represent translation while Euler angle, quaternion, and so(3) spaces were used to express rotations. As mentioned above, if even just one component of $x_{f_{0}^{i}f_{0}^{i+1}}$ is small, the resulting interpolation factor becomes numerically sensitive. In particular, if a front-view camera is mounted on a mobile robot or a vehicle, the motion in the z-axis which is the direction of the camera light rays becomes dominant, meaning that the changes in the x and y-axis will be extremely small. Such conditions have a high possibility of resulting in the aforementioned singular case.

\section{Pose Correction Algorithm}
\label{pose correction algorithm}

\begin{figure}[t!]
\centering
\includegraphics[width=90mm]{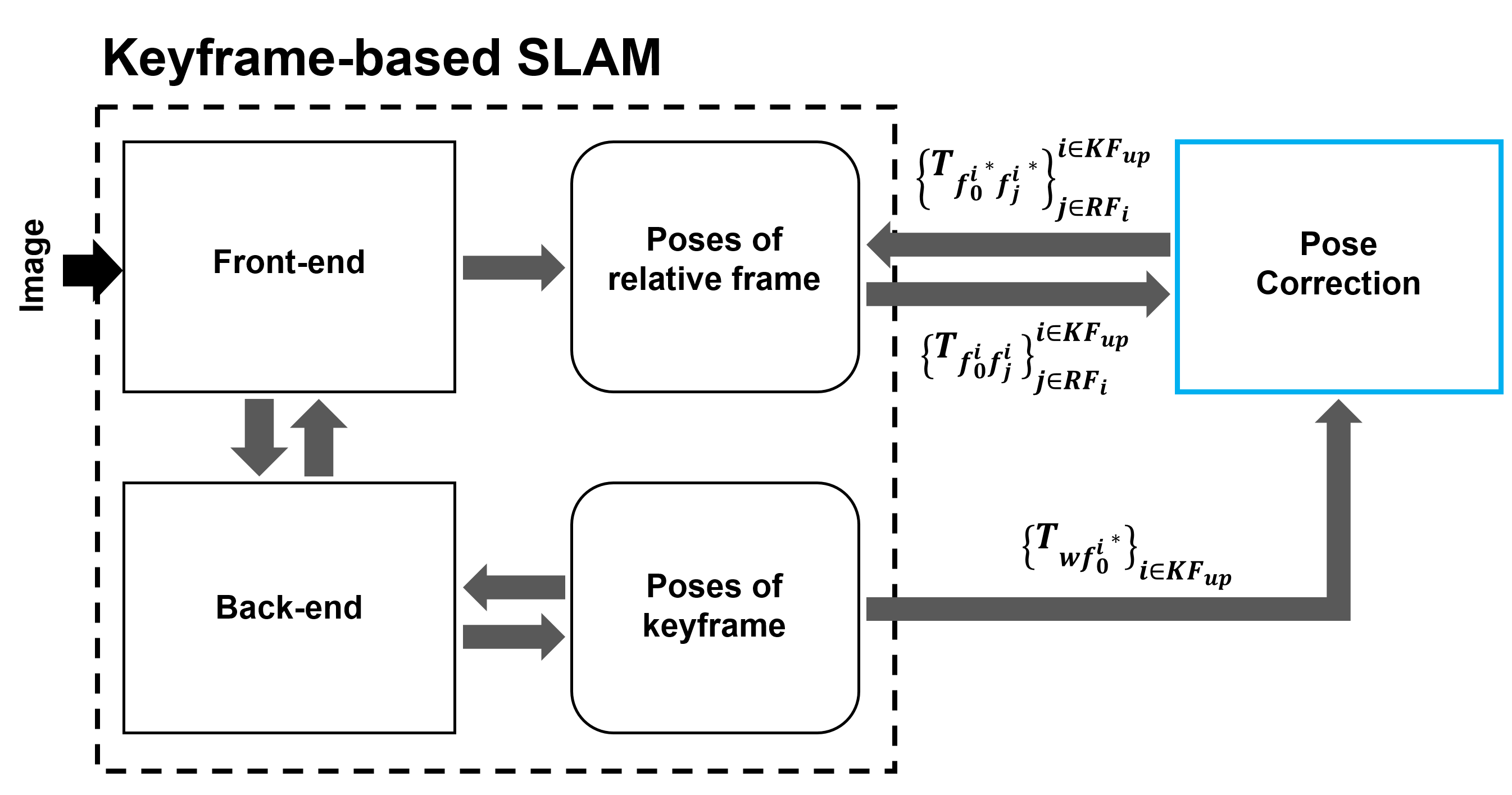} 
\caption{
The overall framework of a keyframe-based SLAM system with the added the pose correction module.
}
\label{fig: framework}
\end{figure}

In this section, the proposed pose correction algorithm is described in detail. The algorithm can be easily integrated into existing keyframe-based SLAM methods as shown in the Fig. \ref{fig: framework}. Typical SLAM front-end systems estimate the relative pose between the newly acquired image and the most relevant keyframe. Back-end systems select keyframes from the input images and perform graph-based optimization with the keyframes as nodes to improve the keyframe poses. Measurement constraints have been typically used to formulate the likelihood function in optimization methods, but not used in non-optimization methods for faster computation. The aim of the proposed algorithm was to preserve the measurement constraints for robustness and accuracy similar to that of the optimization methods, but with a fast computation time similar to that of the non-optimization methods.

The algorithm is triggered when the keyframes are updated and corrects the relative frames connected to the updated keyframes. $\text{KF}_{up}$ in Fig. \ref{fig: framework} represents the index of the keyframe refined by the back-end and $\text{RF}_{i}$ is the relative frame index connected to the $i^{\text{th}}$ keyframe. The correction $T_{f_{j}^{i}{f_{j}^{i}}^{*}}$ of the relative frames $\{f_{j}^{i}\}$ positioned between the $i^{\text{th}}$ and $i+1^{\text{th}}$ keyframes can be computed using the keyframe update information $T_{f_{0}^{i}{f_{0}^{i}}^{*}}$ and $T_{f_{0}^{i+1}{f_{0}^{i+1}}^{*}}$.

\subsection{Measurement Constraints}
\label{measurement constraints}

The correction model will now be derived using the measurement constraints. To simplify the notation, we will omit $f$ from the frame notations $\{f_{j}^{i}\}$ and show $\{i_{j}\}$ when the frame notations are used as subscripts or superscripts. Equations (2) and (3) show the projection equation of $\{f_0^i\}$ and $\{ {f_0^i}^* \}$ respectively.

\begin{align}
    \{ f_0^i \}:& \;\;\;\;\; {}^{i_0}P_k = {}^{i_0}\lambda_k K^{-1} {}^{i_0}\Bar{u}_k \label{eqn: proj1}\\
    \{ {f_0^i}^* \}:& \;\;\;\;\; {}^{{i_0}^*}P_k^* = {}^{{i_0}^*}\lambda_k^* K^{-1} {}^{{i_0}^*}\Bar{u}_k \label{eqn: proj2}
\end{align}
\begin{equation*}
\text{where} \;\;\;\;\;
 {}^{i_0}P_k = \prescript{i_0}{}{\begin{bmatrix}
X \\ Y \\ Z \\
\end{bmatrix}_k}, \;\;\;\; 
{}^{i_0}\Bar{u}_k = \prescript{i_0}{}{\begin{bmatrix}
u \\ v \\ 1 \\
\end{bmatrix}_k}.
\end{equation*}
Here, ${}^{i_0}P_k$, ${}^{i_0}\Bar{u}_k$ and ${}^{i_0}\lambda_k$ are the 3-D position, the homogeneous pixel coordinates, and depth of the $k^{\text{th}}$ feature with respect to $\{f_0^i\}$, respectively. $K$ is the intrinsic parameter matrix, and the position of the updated landmark ${}^{{i_0}^*}P_k$ can be expressed using (\ref{eqn: proj1}) and (\ref{eqn: proj2}) as follows:
\begin{align}
    {}^{{i_0}^*}P^*_k &= R_{{i_0}^*i_0} \left( {}^{i_0}P_k + {}^{i_0}\delta P_k \right) + t_{{i_0}^*i_0} \nonumber \\
    &\approx \frac{{}^{{i_0}^*}\lambda^*_k}{{}^{i_0}\lambda_k} \; {}^{i_0}P_k \;\;\; \left( \because \;\;{}^{i_0}\Bar{u}_k = {}^{{i_0}^*}\Bar{u}_k \right) \label{eqn: landmark}
\end{align}
where $R_{{i_0}^*i_0}$ and $t_{{i_0}^*i_0}$ are the rotation and translation from $\{ {f_0^i}^*\}$ to $\{ f_0^i \}$, respectively. ${}^{i_0}\delta P_k$ is the variation of the $k^{\text{th}}$ landmark position by the refinement process of SLAM. (\ref{eqn: landmark}) was derived using the condition that the measurement of the landmark remains constant regardless of the pose corrections.
The projection equation of $\{f_j^i\}$ and $\{ {f_j^i}^* \}$ with respect to the updated landmark ${}^{{i_0}^*}P^*_k$ shown in (\ref{eqn: landmark}) can now be expressed as follows:
\begin{align}
    \{ f_j^i \}:& \frac{{}^{i_0}\lambda_k}{{}^{{i_0}^*}\lambda^*_k} \; R_{i_ji_0} {}^{{i_0}^*}P^*_k + t_{i_ji_0} = {}^{i_j}\lambda_k K^{-1} {}^{i_j}\Bar{u}_k \label{eqn: proj3}\\
    \{ {f_j^i}^* \}:& R_{{i_j}^*{i_0}^*} {}^{{i_0}^*}P^*_k + t_{{i_j}^*{i_0}^*} = {}^{{i_j}^*}\lambda^*_k K^{-1} {}^{{i_j}^*}\Bar{u}_k \label{eqn: proj4}    
\end{align}
\begin{align}
    {}^{{i_0}^*}P^*_k 
    &= \frac{{}^{{i_0}^*}\lambda^*_k {}^{i_j}\lambda_k}{{}^{i_0}\lambda_k} \; R_{i_0i_j} K^{-1} {}^{i_j}\Bar{u}_k + \frac{{}^{{i_0}^*}\lambda_k^*}{{}^{i_0}\lambda_k} \; t_{i_0i_j} \label{eqn: proj5}\\
    &= {}^{{i_j}^*}\lambda_k^* \; R_{{i_0}^*{i_j}^*} K^{-1} {}^{{i_j}^*}\Bar{u}_k + t_{{i_0}^*{i_j}^*} ~. \label{eqn: proj6}
\end{align} 
Using the fact that the measurement ${}^{i_j}\Bar{u}_k$ observed in each image remains constant regardless of the update, (\ref{eqn: identical1}) can be derived from (\ref{eqn: proj5}) and (\ref{eqn: proj6}).
\begin{align}
&\left(\frac{{}^{{i_0}^*}\lambda^*_k {}^{i_j}\lambda_k}{{}^{{i_j}^*}\lambda^*_k {}^{i_0}\lambda_k} R_{i_0i_j} - R_{{i_0}^*{i_j}^*}\right) {}^{{i_j}^*}\lambda^*_k K^{-1} {}^{{i_j}^*}\Bar{u}_k + \frac{{}^{{i_0}^*}\lambda^*_k}{{}^{i_0}\lambda_k}  t_{i_0i_j} -  t_{{i_0}^*{i_j}^*} = 0 ~.
\label{eqn: identical1}
\end{align}

The depth value of each feature increases as the translational difference between the keyframes in which the features were observed increases. Using this characteristic and assuming that the translation ratio and the depth ratio are equal, the following condition is derived:
\begin{align}
     s_i = \frac{{}^{{i_0}^*}\lambda^*_k}{{}^{i_0}\lambda_k} \approx \frac{\left\lVert t_{{i_0}^*{{(i+1)}_0}^*} \right\rVert^2_2}{\left\lVert t_{{i_0}{{(i+1)}_0}} \right\rVert^2_2}, \;\;\; \frac{{}^{{i_0}^*}\lambda^*_k {}^{i_j}\lambda_k}{{}^{{i_j}^*}\lambda^*_k {}^{i_0}\lambda_k} \approx 1 ~. \label{eqn: identical2}
\end{align}
Applying the (\ref{eqn: identical2}) to (\ref{eqn: identical1}) yields an identical equation (\ref{eqn: identical3}) for $\left( {}^{{i_j}^*}\lambda^*_k K^{-1} {}^{{i_j}^*}\Bar{u}_k \right)$. For the identical equation to hold for all measurements, the solution must be expressed as in (\ref{eqn: identical4}).
\begin{align}
&\left( R_{i_0i_j} - R_{{i_0}^*{i_j}^*} \right) {}^{{i_j}^*}\lambda^*_k K^{-1} {}^{{i_j}^*}\Bar{u}_k + s_i \; t_{i_0i_j} -  t_{{i_0}^*{i_j}^*} = 0 
\label{eqn: identical3}
\end{align}
\begin{align}
 R_{{i_0}^*{i_j}^*} = R_{i_0i_j}, \;\;\;\;  t_{{i_0}^*{i_j}^*} = s_i \; t_{i_0i_j} ~. \label{eqn: identical4}
\end{align}
(\ref{eqn: identical4}) was derived using the measurement constraint between the $i^{\text{th}}$ keyframe and $\{f_j^i\}$. Applying the same procedure to the $i+1^{\text{th}}$ keyframe yields (\ref{eqn: identical5}).
\begin{align}
 R_{{{(i+1)}_0}^*{i_j}^*} = R_{{(i+1)}_0i_j}, \;\;\;\;  t_{{{(i+1)}_0}^*{i_j}^*} = s_i \; t_{{(i+1)}_0i_j} ~. \label{eqn: identical5}
\end{align}

\subsection{Fusion with Two Constraints}
\label{fusion with two constraints}

By fusing the conditions (\ref{eqn: identical4}) and (\ref{eqn: identical5}) derived previously, $T_{i_{0}{i_{j}}^{*}}$ can now be computed. The gap between the solutions to the aforementioned conditions can be expressed as follows:
\begin{align}
 \delta R =& \;{R_{{i_0}^*{i_j}^*}^{KF_i}}^T R_{{i_0}^*{{(i+1)}_0}^*} R_{{{(i+1)}_0}^*{i_j}^*}^{KF_{i+1}} \label{eqn: fusion1} \\
 =& \; {R_{i_0i_j}}^T R_{{i_0}^*{{(i+1)}_0}^*} R_{{(i+1)}_0i_j} \nonumber \\
 \delta t =& \; t_{{i_0}^*{i_j}^*}^{KF_i} + R_{{i_j}^*{i_0}^*} t_{{i_0}^*{{(i+1)}_0}^*} + R_{{i_j}^*{{(i+1)}_0}^*} t_{{{(i+1)}_0}^*{i_j}^*}^{KF_{i+1}}  \nonumber \\
 =& \; R_{{i_j}^*{i_0}^*} \left( t_{{i_0}^*{{(i+1)}_0}^*} - s_i (t_{i_0i_j} - R_{{i_0}^*{{(i+1)}_0}^*} t_{{{(i+1)}_0}{i_0}}) \right) \nonumber
\end{align}
where $R_{{i_0}^*{i_j}^*}^{KF_i}$ and $t_{{i_0}^*{i_j}^*}^{KF_i}$ are the corrected relative rotation and translation computed from (\ref{eqn: identical4}), and $R_{{i_0}^*{i_j}^*}^{KF_{i+1}}$ and $t_{{i_0}^*{i_j}^*}^{KF_{i+1}}$ are the correction terms computed from (\ref{eqn: identical5}). The $\delta R$ and $\delta t$ terms in (\ref{eqn: fusion1}) are expressed with respect to $\{ {f_j^i}^* \}$, which is estimated under the conditions given in (\ref{eqn: identical4}). To compensate for the gap, fusion as expressed in (\ref{eqn: fusion2}) is performed to estimate the corrected relative frame.
\begin{align}
    R_{{i_0}^*{i_j}^*} &= R_{{i_0}^*{i_j}^*}^{KF_i} \cdot \text{SLERP} \left( \delta R, \; \alpha^i_j \right) \\
    t_{{i_0}^*{i_j}^*} &= t_{{i_0}^*{i_j}^*}^{KF_i} + \text{LERP} \left( R_{{i_0}^*{i_j}^*} \delta t, \; \alpha^i_j \right) \label{eqn: fusion2}
\end{align}
where SLERP($\cdot$) and LERP($\cdot$) are spherical linear interpolation and linear interpolation functions, respectively. $\delta R$ is converted to a quaternion space to be utilized in the SLERP function. $\alpha^i_j$ is the interpolation factor which should reflect the reliability of conditions given by (\ref{eqn: identical4}) and (\ref{eqn: identical5}). Since the number of reliable edges increases as the distance between frames decreases due to the increase in the number of shared features, the ratio of the distance from $\{ {f_j^i} \}$ to $\{ {f_0^i} \}$ and the distance from $\{ {f_j^i} \}$ to $\{ {f_0^{i+1}} \}$ was used as the interpolation factor in this paper.

\section{Experimental Result}
\label{experimental result}

This section provides the results of the proposed pose correction algorithm integrated with ORB-SLAM2 \cite{mur2017orb} which is one of the most popular keyframe-based SLAM. As mentioned above, the existing interpolation-based methods were tested in various vector spaces to function as a baseline for comparison. The interpolations for translation were done in XYZ and the translation component $v$ of the se(3) spaces, while the rotation components were interpolated in Euler angles, quaternion, and so(3) spaces. The stereo images of KITTI \cite{geiger2013vision} and EuRoC \cite{burri2016euroc} benchmarks datasets were used for analysis. ORB-SLAM2 was used to generate the poses of frames, though any appropriate keyframe-based SLAM can be applied. 

There are two types of refinements that occur in the back-end of SLAM: local BA and global BA. Local BA occurs when a new keyframe is added and refines only keyframes that have a strong connection to the newly added keyframe. Global BA occurs when a loop is detected and refines all keyframes that are in the map. The proposed pose correction module is triggered whenever local or global BA takes place and uses the updated keyframes and their relative frames as inputs. We analyze the accuracy of the corrected relative frame poses and the computation time required to compute the correction. However, because the keyframe poses computed by the SLAM system inherently contain error, the difference between the corrected relative frames poses and the ground truth (GT) may not purely reflect the correction performance. Therefore, an additional post processing step was introduced to the SLAM system to directly evaluate the correction performance of the algorithm. The poses of keyframes that lie on the estimated trajectory by ORB-SLAM2 with the correction module was additionally updated to their GT poses so that the keyframes now lie on the GT. The proposed algorithm was used to correct the relative frames so that the final output of the SLAM is corrected to the GT.  The difference between these final corrected relative frame poses and the GT poses was used as the error metric for correction. Tests were performed on a laptop (Y520-15IKBN, 16GB RAM with Intel i7-7700HQ @ 2.80GHz $\times$ 4cores).

\subsection{KITTI Dataset}

The KITTI dataset \cite{geiger2013vision} is generated from a stereo camera mounted on top of a vehicle, where the yaw motion and the camera z-axis movement are dominant with minimal motion along other axis due to the characteristics of a vehicle platform. 

The ORB-SLAM2 and correction algorithm results obtained from the sequences (00-10) of the KITTI dataset are summarized in table \ref{tab: SLAMkittiresult} and \ref{tab: kittiresult} respectively. No-correction in table \ref{tab: kittiresult} refers to the results obtained from the simple concatenation of relative frames and keyframes without correction. The proposed algorithm outperformed all the baseline interpolation methods in all sequences except for sequence 01. As can be seen in table \ref{tab: SLAMkittiresult}, almost all image frames became keyframes using ORB-SLAM2 in sequence 01, meaning that the correction module had a minimal effect. For translation, the proposed algorithm nearly doubled the mean accuracy of the baseline method in sequences 00, 02, 03, 07 and 08. Furthermore, the algorithm yielded a lower standard deviation when compared to the baseline methods. Since standard deviation indicates the robustness of the system in a variety of situations, it can be concluded that the proposed algorithm is not numerically sensitive compared to the baseline methods. Rotation, on the other hand, does not exhibit significant differences in accuracy between methods. This is because the KITTI dataset was acquired using a ground vehicle, resulting in very little rotation aside from yaw. There are, however, significant differences in the standard deviation for rotation, meaning that singular cases occur in certain areas of the sequences, resulting in significant error. In some sequences, especially for rotations, the baseline methods yielded worse results than the no-correction method. Because rotation has such a small error even prior to correction, the numerical error has a significant effect on the results.

\begin{figure}[]
\centering
\includegraphics[width=120mm]{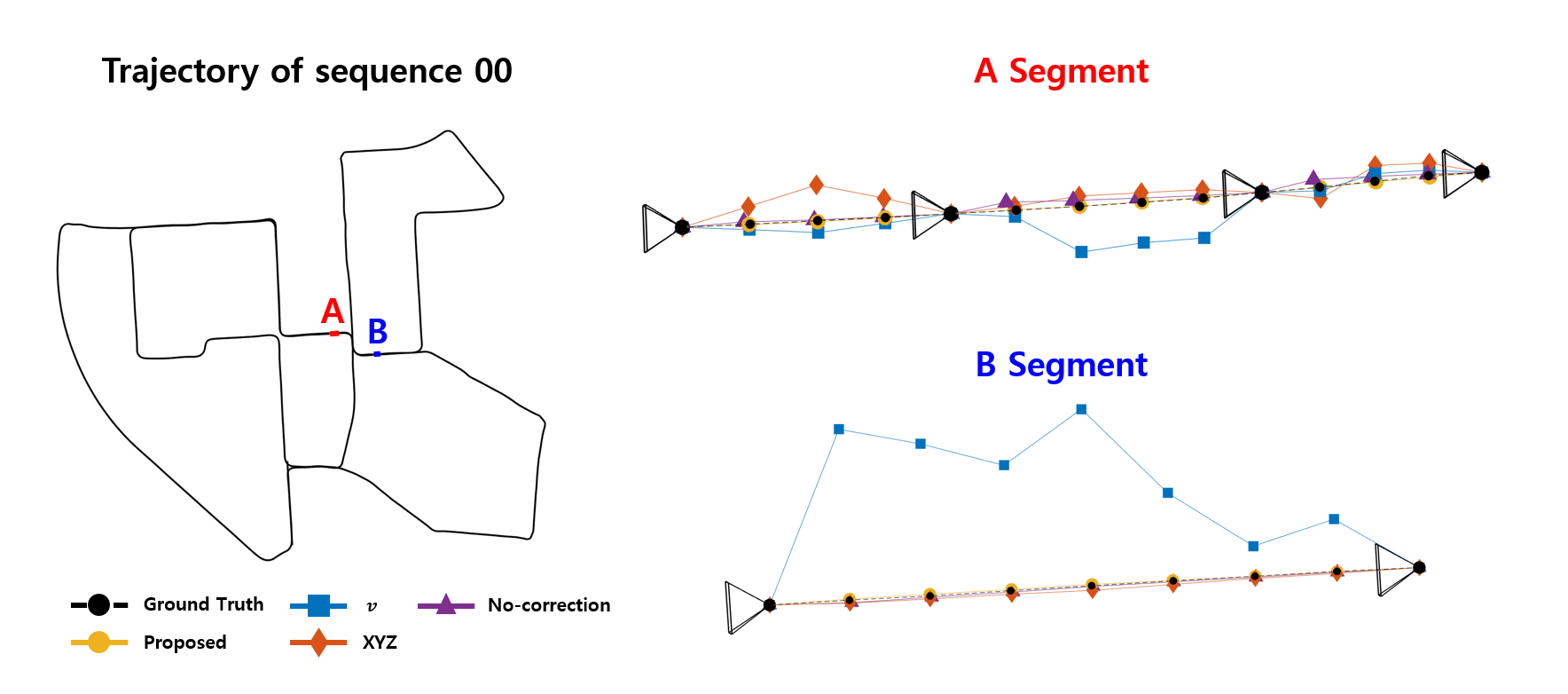} 
\caption{
The translational results of the correction algorithm for two select segments in sequence 00 of KITTI.
}
\label{fig:kitti}
\end{figure}

The resultant trajectory from each translation space for the select segments A and B in sequence 00 is shown in figure \ref{fig:kitti}. Segment A visualizes the varying performance of the baseline methods, as both the XYZ and $v$ interpolations stray wildly from the GT poses, even more so than the no-correction method. The proposed method, however, was able to remain consistent with the GT poses throughout the entire segment. Segment B shows the singular case in $v$, resulting in a huge deviation away from the GT. It is worth noting that the singularity occurred only in the $v$ space interpolation and not the XYZ space interpolation. This is a clear depiction of the numerical sensitivity of the existing baseline methods, as such deviations can result in large error that may be even worse than the cases without correction at all. The proposed algorithm, however, performed robustly in both cases, demonstrating the improved accuracy and numerical robustness the algorithm has over the baseline methods.



\subsection{EuRoC Dataset}

The EuRoC dataset \cite{burri2016euroc} is generated from a stereo camera mounted on a micro aerial vehicle (MAV) and contains a more diverse range of motion compared to the KITTI data. The MAV was flown in an industrial environment (machine room) and two different rooms with a motion capture system in place. There are a total of 11 sequences, with each sequence classified as easy, medium, and difficult depending on the motion of the MAV and the room environment. Since ORB-SLAM2 does not provide sufficient results for V2\textunderscore 03\textunderscore difficult due to significant motion blur in some of the frames, the particular sequences were not used in this paper.

The results of the ORB-SLAM2 and correction algorithm obtained from the remaining ten sequences are described in tables \ref{tab: SLAMeurocresult} and \ref{tab: eurocresult}, respectively. The EuRoC dataset is generated in a much smaller environment than the KITTI dataset, resulting in sparser keyframes. Furthermore, because the motion of a MAV is erratic compared to a ground vehicle, the pose error from SLAM is more significant than the KITTI dataset. Hence, the need for a correction algorithm is more apparent. The proposed algorithm demonstrates significantly improved results for both translation and rotation and the improvements are clearer than the KITTI dataset. In sequence V102, for example, the standard deviation of the algorithm was almost four times lower than that of the second best method for translation (no-correction) and five times lower than that of the second best method for rotation (no-correction), demonstrating its robustness. The mean and median values have also been halved, showcasing the accuracy of the algorithm.

\begin{figure}[]
\centering
\includegraphics[width=120mm]{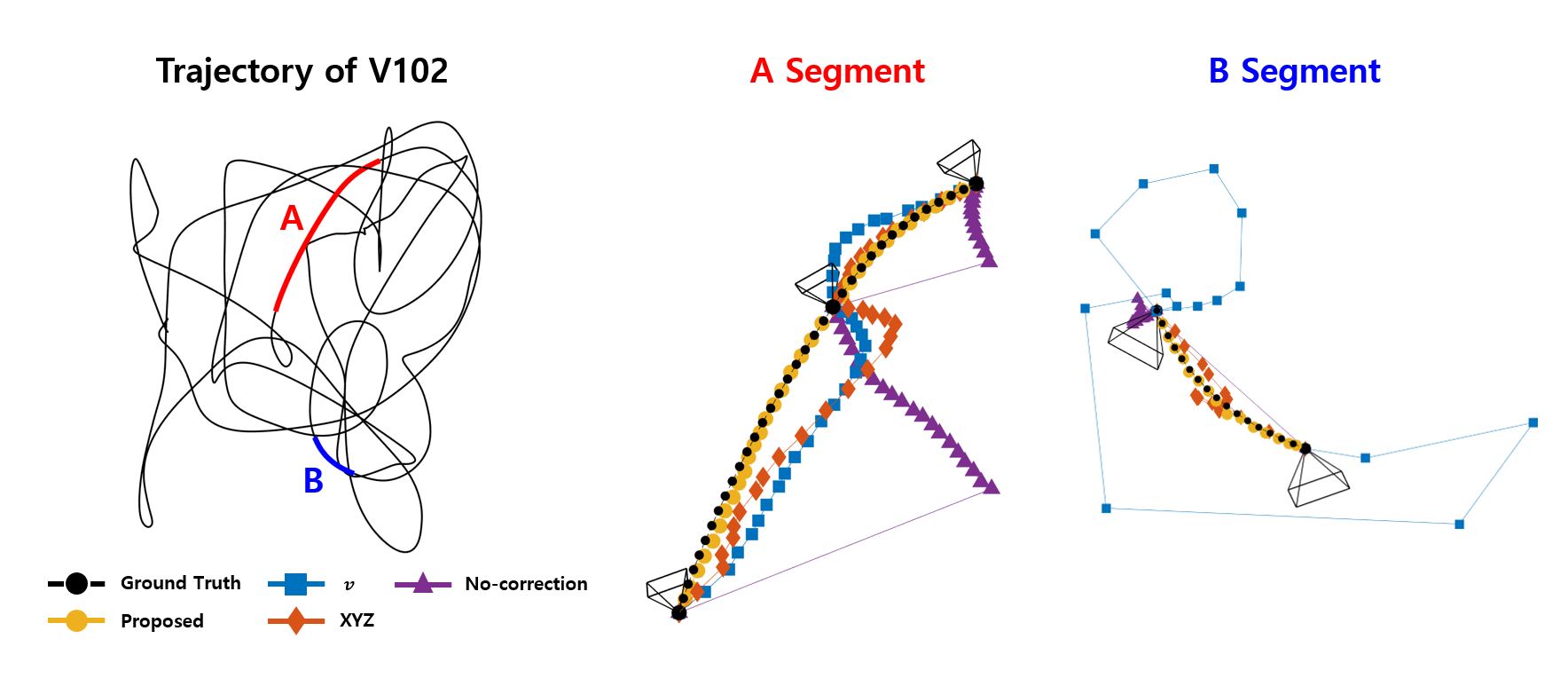} 
\caption{
The results of the corrected translation for two select segments in V102 of EuRoC.}
\label{fig:euroc}
\end{figure}

The corrected trajectory for select segments of sequence V102 is shown in figure \ref{fig:euroc}. Unlike the KITTI dataset, the no-correction method becomes meaningless, as the concatenation of relative frames and keyframes results in discontinuous trajectories as shown in both segments A and B. Furthermore, in both segments, the baseline methods failed to remain consistent with the GT, showing significant deviations in particular around the second keyframe in segment A. In segment B, the numerical sensitivity of the $v$ space interpolation method causes the corrected poses to deviate significantly from the GT trajectory. The XYZ space interpolation method is also unable to achieve the desired correction and results in significant clustering around the midpoint between the two keyframes. The proposed method, on the other hand, was able to remain close to the GT trajectory with no singularities. Hence, even under erratic motion that causes the baseline methods to fail, the proposed algorithm was still able to generate accurate and robust corrections.

\subsection{Computation time}

The proposed correction module triggers whenever keyframe refinement, or in other words BA, occurs in keyframe-based SLAM. Table \ref{tab: comptime} shows the amount of time required for the algorithm to compute a single correction in a MATLAB environment. Although there were no significant differences between the proposed and baseline algorithms for translation, the proposed algorithm required the most time for rotation. This is because the SLERP algorithm used in the proposed algorithm requires more computations than a simple interpolation approach. However, the difference in the median computation time is approximately 1 millisecond and the algorithm only runs when a new keyframe is selected or when a loop is closed, meaning that the computation time required is insignificant when compared to the entire SLAM pipeline. In addition, the standard deviation of the computation time is large compared to the median value. This is due to the presence of loops within certain sequences, which results in a global BA. As discussed previously, global BA updates all keyframes, meaning that all relative frames must be corrected. Hence, typical computation time for a typical local BA is similar to that of the median value.

\begin{table}[h]
\centering
\captionsetup{justification=raggedright,singlelinecheck=false}
\caption{
The computation time taken for a single correction operation. Each cell contains the mean $\pm$ standard deviation, and (median). $v$ represents the translation component of se(3) space.}
\label{tab: comptime}
\begin{tabular}{cccccccc}
\hline\hline
\multicolumn{3}{c}{Translation (msec)}                                                                                                                                                                                                                                                &  & \multicolumn{4}{c}{Rotation (msec)}                                                                                                                                                                                                                                                                                                                                              \\ \cline{1-3} \cline{5-8} 
XYZ                                                                                      & $v$                                                                                        & Proposed                                                                                 &  & Euler                                                                                      & Quat                                                                                     & so(3)                                                                                    & Proposed                                                                                 \\ \hline
\begin{tabular}[c]{@{}c@{}}0.340$\pm$0.86\\ (0.170)\end{tabular} & \begin{tabular}[c]{@{}c@{}}0.698$\pm$1.74\\ (0.313)\end{tabular} & \begin{tabular}[c]{@{}c@{}}0.356$\pm$0.89\\ (0.135)\end{tabular} &  & \begin{tabular}[c]{@{}c@{}}1.781$\pm$4.66\\ (0.702)\end{tabular} & \begin{tabular}[c]{@{}c@{}}2.205$\pm$5.66\\ (0.944)\end{tabular} & \begin{tabular}[c]{@{}c@{}}0.689$\pm$1.69\\ (0.318)\end{tabular} & \begin{tabular}[c]{@{}c@{}}3.916$\pm$9.87\\ (1.441)\end{tabular} \\ \hline\hline
\end{tabular}
\end{table}

\begin{table}[]
\centering
\captionsetup{justification=raggedright,singlelinecheck=false}
\caption{
The results of ORB-SLAM2 in KITTI dataset. The number of keyframes and all frames and loop closures are indicated as shown.
}
\label{tab: SLAMkittiresult}
\setlength{\tabcolsep}{2.5pt}
\begin{tabular}{lccccccccccc}
\hline
\hline
KITTI    & 00   & 01   & 02   & 03  & 04  & 05   & 06   & 07   & 08   & 09   & 10   \\ \hline
\# of Keyframes & 1355 & 1047 & 1742 & 226 & 155 & 717  & 473  & 251  & 1199 & 588  & 321  \\
\# of All Frames & 4541 & 1101 & 4661 & 801 & 271 & 2701 & 1101 & 1101 & 4071 & 1504 & 1201 \\
Loop     & O    & X    & O    & X   & X   & O    & O    & O    & X    & X    & X   \\ \hline \hline
\end{tabular}
\end{table}

\begin{table}[]
\centering
\captionsetup{justification=raggedright,singlelinecheck=false}
\caption{
The summary of results for the KITTI dataset. The blue indicates the lowest error while red indicates the highest error. The conventions are same as in table \ref{tab: comptime}.}
\label{tab: kittiresult}
\resizebox{\textwidth}{!}{
\begin{tabular}{lcccccccccc}
\hline
\hline
                        & \multicolumn{4}{c}{Translation (cm)}                                                                                                                                  &  & \multicolumn{5}{c}{Rotation ($\times 10^{-1}$deg)}                                                                                                                                                                                             \\ \cline{2-5} \cline{7-11} 
Sequence                & No-Correction  & XYZ                                                   & $v$                                                   & Proposed                                                && No-Correction & Euler                                                   & Quat                                                  & so(3)     
 & Proposed                                              \\ \hline
00                      & 
\begin{tabular}[c]{@{}c@{}}2.034$\pm$1.76\\(1.425)\end{tabular} & \begin{tabular}[c]{@{}c@{}}1.919$\pm$3.91\\ (0.984)\end{tabular} & \begin{tabular}[c]{@{}c@{}}\rednote{2.949$\pm$9.84}\\\rednote{ (1.037)}\end{tabular} &
 \begin{tabular}[c]{@{}c@{}}\bluenote{0.947$\pm$0.79}\\\bluenote{(0.698)} \end{tabular} &  &
 \begin{tabular}[c]{@{}c@{}}0.618$\pm$0.59\\(0.445)\end{tabular} & \begin{tabular}[c]{@{}c@{}}0.891$\pm$1.37\\(0.472)\end{tabular} & \begin{tabular}[c]{@{}c@{}}0.954$\pm$1.60\\(0.473)\end{tabular} & \begin{tabular}[c]{@{}c@{}}\rednote{0.955$\pm$1.60}\\\rednote{(0.473)}\end{tabular} & 
\begin{tabular}[c]{@{}c@{}}\bluenote{0.473$\pm$0.35}\\\bluenote{(0.378)}\end{tabular} \\
01                      & 
\begin{tabular}[c]{@{}c@{}}\rednote{1.874$\pm$0.52}\\\rednote{(1.890)}\end{tabular} & 
\begin{tabular}[c]{@{}c@{}}\bluenote{0.885$\pm$0.46}\\\bluenote{(0.788)}\end{tabular} & 
\begin{tabular}[c]{@{}c@{}}1.083$\pm$0.84\\(0.876)\end{tabular} & 
\begin{tabular}[c]{@{}c@{}}0.915$\pm$0.46\\(0.833)\end{tabular} &  & 
\begin{tabular}[c]{@{}c@{}}0.190$\pm$0.08\\(0.188)\end{tabular} & 
\begin{tabular}[c]{@{}c@{}}0.190$\pm$0.08\\ (0.188)\end{tabular} & 
\begin{tabular}[c]{@{}c@{}}\rednote{0.191$\pm$0.09}\\\rednote{(0.189)}\end{tabular} & \begin{tabular}[c]{@{}c@{}}\rednote{0.191$\pm$0.09}\\\rednote{(0.189)}\end{tabular} & 
\begin{tabular}[c]{@{}c@{}}\bluenote{0.188$\pm$0.08}\\\bluenote{(0.188)}\end{tabular} \\
02                      & 
\begin{tabular}[c]{@{}c@{}}1.737$\pm$1.13\\(1.458)\end{tabular} & \begin{tabular}[c]{@{}c@{}}1.535$\pm$1.90\\(0.993)\end{tabular} & \begin{tabular}[c]{@{}c@{}}\rednote{1.786$\pm$2.37}\\\rednote{(1.05)}\end{tabular} &
\begin{tabular}[c]{@{}c@{}}\bluenote{0.916$\pm$0.61}\\\bluenote{(0.762)}\end{tabular} &  & \begin{tabular}[c]{@{}c@{}}0.442$\pm$0.31\\(0.362)\end{tabular} & \begin{tabular}[c]{@{}c@{}}0.561$\pm$0.56\\(0.377)\end{tabular} & \begin{tabular}[c]{@{}c@{}}0.591$\pm$0.67\\ (0.382)\end{tabular} & 
\begin{tabular}[c]{@{}c@{}}\rednote{0.592$\pm$0.67}\\\rednote{(0.383)}\end{tabular} &
\begin{tabular}[c]{@{}c@{}}\bluenote{0.392$\pm$0.25}\\ \bluenote{(0.327)}\end{tabular} \\
03                      & 
\begin{tabular}[c]{@{}c@{}}1.455$\pm$0.87\\ (1.228)\end{tabular} & \begin{tabular}[c]{@{}c@{}}1.610$\pm$2.07\\ (0.985)\end{tabular} &
\begin{tabular}[c]{@{}c@{}}\rednote{1.692$\pm$2.10}\\\rednote{(1.002)}\end{tabular} & \begin{tabular}[c]{@{}c@{}}\bluenote{0.775$\pm$0.47}\\\bluenote{ (0.665)}\end{tabular} &  & \begin{tabular}[c]{@{}c@{}}0.478$\pm$0.23\\ (0.403)\end{tabular} & \begin{tabular}[c]{@{}c@{}}\rednote{0.594$\pm$0.53}\\\rednote{ (0.417)}\end{tabular} & \begin{tabular}[c]{@{}c@{}}0.558$\pm$0.40\\ (0.417)\end{tabular} &
\begin{tabular}[c]{@{}c@{}}0.559$\pm$0.40\\(0.417)\end{tabular} & \begin{tabular}[c]{@{}c@{}}\bluenote{0.456$\pm$0.21}\\\bluenote{(0.393)}\end{tabular} \\
04                      & \begin{tabular}[c]{@{}c@{}}\rednote{1.045$\pm$0.50}\\\rednote{(1.021)}\end{tabular} & \begin{tabular}[c]{@{}c@{}}0.798$\pm$0.50\\ (0.701)\end{tabular} &
\begin{tabular}[c]{@{}c@{}}0.818$\pm$0.52\\(0.731)\end{tabular} & \begin{tabular}[c]{@{}c@{}}\bluenote{0.726$\pm$0.42}\\ \bluenote{(0.658)}\end{tabular} &  & \begin{tabular}[c]{@{}c@{}}0.267$\pm$0.15\\(0.023)\end{tabular} & \begin{tabular}[c]{@{}c@{}}\rednote{0.346$\pm$0.32}\\\rednote{ (0.242)}\end{tabular} & \begin{tabular}[c]{@{}c@{}}0.346$\pm$0.32\\ (0.240)\end{tabular} &
\begin{tabular}[c]{@{}c@{}}0.346$\pm$0.32\\(0.239)\end{tabular} & \begin{tabular}[c]{@{}c@{}}\bluenote{0.254$\pm$0.14}\\ \bluenote{(0.225)}\end{tabular} \\
05                      &
\begin{tabular}[c]{@{}c@{}}\rednote{1.495$\pm$1.32}\\\rednote{ (1.074)}\end{tabular} & \begin{tabular}[c]{@{}c@{}}1.008$\pm$0.83\\ (0.747)\end{tabular} &
\begin{tabular}[c]{@{}c@{}}1.166$\pm$1.31\\(0.757)\end{tabular} & \begin{tabular}[c]{@{}c@{}}\bluenote{0.849$\pm$0.63}\\\bluenote{ (0.656)}\end{tabular} &  & \begin{tabular}[c]{@{}c@{}}0.469$\pm$0.34\\ (0.375)\end{tabular} & \begin{tabular}[c]{@{}c@{}}0.672$\pm$0.85\\ (0.420)\end{tabular} & \begin{tabular}[c]{@{}c@{}}\rednote{0.746$\pm$1.14}\\\rednote{(0.422)}\end{tabular} &
\begin{tabular}[c]{@{}c@{}}\rednote{0.746$\pm$1.14}\\\rednote{(0.422)}\end{tabular} & \begin{tabular}[c]{@{}c@{}}\bluenote{0.405$\pm$0.28}\\\bluenote{(0.328)}\end{tabular} \\
06                      & 
\begin{tabular}[c]{@{}c@{}}\rednote{1.204$\pm$1.06}\\ \rednote{(0.855)}\end{tabular} & \begin{tabular}[c]{@{}c@{}}0.813$\pm$0.81\\ (0.591)\end{tabular} & 
\begin{tabular}[c]{@{}c@{}}0.814$\pm$0.80\\(0.585)\end{tabular} & \begin{tabular}[c]{@{}c@{}}\bluenote{0.639$\pm$0.43}\\ \bluenote{(0.509)}\end{tabular} &  & \begin{tabular}[c]{@{}c@{}}0.339$\pm$0.25\\ (0.258)\end{tabular} & \begin{tabular}[c]{@{}c@{}}0.711$\pm$1.78\\(0.292)\end{tabular} & \begin{tabular}[c]{@{}c@{}}0.658$\pm$1.47\\ (0.292)\end{tabular} &   
\begin{tabular}[c]{@{}c@{}}\rednote{0.723$\pm$1.97}\\\rednote{(0.295)}\end{tabular} & \begin{tabular}[c]{@{}c@{}}\bluenote{0.283$\pm$0.17}\\\bluenote{ (0.248)}\end{tabular} \\
07                      & 
\begin{tabular}[c]{@{}c@{}}2.034$\pm$2.17\\(1.267)\end{tabular} & \begin{tabular}[c]{@{}c@{}}2.648$\pm$4.74\\ (0.875)\end{tabular} &
\begin{tabular}[c]{@{}c@{}}\rednote{2.784$\pm$5.02}\\\rednote{(0.886)}\end{tabular} & \begin{tabular}[c]{@{}c@{}}\bluenote{1.372$\pm$2.12}\\\bluenote{(0.701)}\end{tabular} &  & \begin{tabular}[c]{@{}c@{}}0.537$\pm$0.42\\ (0.428)\end{tabular} & \begin{tabular}[c]{@{}c@{}}\rednote{0.717$\pm$1.03}\\\rednote{ (0.423)}\end{tabular} & \begin{tabular}[c]{@{}c@{}}0.679$\pm$0.84\\(0.423)\end{tabular} &
\begin{tabular}[c]{@{}c@{}}0.679$\pm$0.84\\(0.423)\end{tabular} & \begin{tabular}[c]{@{}c@{}}\bluenote{0.418$\pm$0.28}\\\bluenote{ (0.342)}\end{tabular} \\
08                      & 
\begin{tabular}[c]{@{}c@{}}\rednote{2.153$\pm$2.01}\\ \rednote{(1.49)}\end{tabular} & \begin{tabular}[c]{@{}c@{}}1.464$\pm$1.72\\(0.925)\end{tabular} &
\begin{tabular}[c]{@{}c@{}}1.571$\pm$1.89\\(0.957)\end{tabular} & \begin{tabular}[c]{@{}c@{}}\bluenote{0.883$\pm$0.62}\\\bluenote{(0.714)}\end{tabular} &  & \begin{tabular}[c]{@{}c@{}}0.481$\pm$0.35\\ (0.394)\end{tabular} & \begin{tabular}[c]{@{}c@{}}0.658$\pm$0.74\\ (0.420)\end{tabular} & \begin{tabular}[c]{@{}c@{}}0.759$\pm$1.13\\(0.425)\end{tabular} &
\begin{tabular}[c]{@{}c@{}}\rednote{0.762$\pm$1.14}\\\rednote{(0.425)}\end{tabular} & \begin{tabular}[c]{@{}c@{}}\bluenote{0.394$\pm$0.26}\\\bluenote{ (0.335)}\end{tabular} \\
09                      & 
\begin{tabular}[c]{@{}c@{}}\rednote{1.362$\pm$0.82}\\\rednote{ (1.167)}\end{tabular} & \begin{tabular}[c]{@{}c@{}}0.997$\pm$0.70\\(0.819)\end{tabular} &
\begin{tabular}[c]{@{}c@{}}1.179$\pm$1.12\\(0.875)\end{tabular} & \begin{tabular}[c]{@{}c@{}}\bluenote{0.868$\pm$0.52}\\\bluenote{(0.755)}\end{tabular} &  & \begin{tabular}[c]{@{}c@{}}0.415$\pm$0.27\\ (0.367)\end{tabular} & \begin{tabular}[c]{@{}c@{}}0.499$\pm$0.44\\ (0.378)\end{tabular} & \begin{tabular}[c]{@{}c@{}}0.527$\pm$0.53\\(0.382)\end{tabular} & \begin{tabular}[c]{@{}c@{}}\rednote{0.527$\pm$0.53}\\\rednote{(0.383)}\end{tabular} &
\begin{tabular}[c]{@{}c@{}}\bluenote{0.381$\pm$0.24}\\\bluenote{ (0.334)}\end{tabular} \\
10                      & 
\begin{tabular}[c]{@{}c@{}}\rednote{1.484$\pm$1.22}\\\rednote{ (1.108)}\end{tabular} & \begin{tabular}[c]{@{}c@{}}1.078$\pm$0.98\\(0.788)\end{tabular} & 
\begin{tabular}[c]{@{}c@{}}1.443$\pm$1.77\\(0.854)\end{tabular} &
\begin{tabular}[c]{@{}c@{}}\bluenote{0.806$\pm$0.62}\\\bluenote{(0.626)}\end{tabular} &  & \begin{tabular}[c]{@{}c@{}}0.502$\pm$0.32\\ (0.447)\end{tabular} & \begin{tabular}[c]{@{}c@{}}0.719$\pm$0.71\\ (0.536)\end{tabular} & \begin{tabular}[c]{@{}c@{}}0.730$\pm$0.76\\(0.536)\end{tabular} &
\begin{tabular}[c]{@{}c@{}}\rednote{0.730$\pm$0.76}\\\rednote{(0.537)}\end{tabular} & \begin{tabular}[c]{@{}c@{}}\bluenote{0.438$\pm$0.28}\\\bluenote{ (0.386)}\end{tabular} \\ \hline\hline
\end{tabular}}
\end{table}

\begin{table}[]
\centering
\captionsetup{justification=raggedright,singlelinecheck=false}
\caption{
The number of keyframes and all frames within the EuRoC dataset. The sequences with loop closures are indicated.
}
\label{tab: SLAMeurocresult}
\setlength{\tabcolsep}{2.5pt}
\begin{tabular}{lccccccccccc}
\hline \hline
\multirow{2}{*}{EuRoC} & \multicolumn{5}{c}{Machine Hall} &  & \multicolumn{5}{c}{Vicon Room}   \\ \cline{2-6} \cline{8-12} 
                       & 01   & 02   & 03   & 04   & 05   &  & 101  & 102  & 103  & 201  & 202  \\ \hline
\# of Keyframes                     & 483  & 431  & 436  & 302  & 353  &  & 109  & 151  & 208  & 216  & 271  \\
\# of All Frames                     & 3638 & 2999 & 2662 & 1976 & 2221 &  & 2871 & 1670 & 2093 & 2148 & 2309 \\
Loop                   & X    & X    & X    & X    & O    &  & X    & X    & O    & X    & X    \\ \hline \hline
\end{tabular}
\end{table}

\begin{table}[]
\centering
\captionsetup{justification=raggedright,singlelinecheck=false}
\caption{
The summary of results for the EuRoC dataset. The conventions are same as in table \ref{tab: comptime}.}
\label{tab: eurocresult}
\resizebox{\textwidth}{!}{
\begin{tabular}{lcccccccccc}
\hline
\hline
                        & \multicolumn{4}{c}{Translation (cm)}                                                                                                                                  &  & \multicolumn{5}{c}{Rotation (deg)}                                                                                                                                                                                             \\ \cline{2-5} \cline{7-11} 
Sequence              & No-Correction  & XYZ                                                   & $v$                                                     & Proposed                                              &  & No-Correction & Euler                                                   & Quat                                                  & so(3)                                                 & Proposed                                              \\ \hline
MH01                   & 
\begin{tabular}[c]{@{}c@{}}\rednote{8.130$\pm$7.53}\\\rednote{(4.817)}\end{tabular} &
\begin{tabular}[c]{@{}c@{}}4.385$\pm$9.58\\ (1.076)\end{tabular} &
\begin{tabular}[c]{@{}c@{}}5.026$\pm$10.84\\(1.219)\end{tabular} & \begin{tabular}[c]{@{}c@{}}\bluenote{1.937$\pm$2.28}\\\bluenote{(1.076)}\end{tabular} &  & \begin{tabular}[c]{@{}c@{}}\rednote{5.904$\pm$7.69}\\\rednote{(2.095)}\end{tabular} & \begin{tabular}[c]{@{}c@{}}4.014$\pm$5.14\\ (1.453)\end{tabular} & \begin{tabular}[c]{@{}c@{}}3.827$\pm$4.68\\ (1.529)\end{tabular} &
\begin{tabular}[c]{@{}c@{}}3.840$\pm$4.70\\ (1.532)\end{tabular} &\begin{tabular}[c]{@{}c@{}}\bluenote{3.214$\pm$4.78}\\\bluenote{(1.110)}\end{tabular} \\
MH02                        & 
\begin{tabular}[c]{@{}c@{}}\rednote{7.592$\pm$7.90}\\\rednote{(4.783)}\end{tabular} & \begin{tabular}[c]{@{}c@{}}3.028$\pm$5.03\\ (0.913)\end{tabular} &
\begin{tabular}[c]{@{}c@{}}2.947$\pm$4.87\\ (0.977)\end{tabular} & \begin{tabular}[c]{@{}c@{}}\bluenote{1.974$\pm$2.75}\\\bluenote{(0.766)}\end{tabular} &  & \begin{tabular}[c]{@{}c@{}}2.212$\pm$2.70\\ (1.174)\end{tabular} & \begin{tabular}[c]{@{}c@{}}2.456$\pm$4.20\\ (0.724)\end{tabular} &
\begin{tabular}[c]{@{}c@{}}2.632$\pm$4.73\\ (0.706)\end{tabular} & \begin{tabular}[c]{@{}c@{}}\rednote{2.654$\pm$4.79}\\\rednote{(0.706)}\end{tabular} & \begin{tabular}[c]{@{}c@{}}\bluenote{1.672$\pm$2.17}\\\bluenote{(0.759)}\end{tabular} \\
MH03                       & 
\begin{tabular}[c]{@{}c@{}}\rednote{22.693$\pm$24.15}\\ \rednote{(15.429)}\end{tabular} &
\begin{tabular}[c]{@{}c@{}}11.755$\pm$16.50\\ (4.596)\end{tabular} & \begin{tabular}[c]{@{}c@{}}15.135$\pm$24.25\\(5.254)\end{tabular} & \begin{tabular}[c]{@{}c@{}}\bluenote{5.047$\pm$6.31}\\\bluenote{(2.244)}\end{tabular} &  & \begin{tabular}[c]{@{}c@{}}\rednote{4.455$\pm$5.20}\\\rednote{(2.852)}\end{tabular} & \begin{tabular}[c]{@{}c@{}}3.687$\pm$5.12\\ (1.502)\end{tabular} &
\begin{tabular}[c]{@{}c@{}}4.028$\pm$6.64\\ (1.479)\end{tabular} & \begin{tabular}[c]{@{}c@{}}4.247$\pm$7.55\\(1.496)\end{tabular} & \begin{tabular}[c]{@{}c@{}}\bluenote{3.236$\pm$3.79}\\\bluenote{(1.374)}\end{tabular} \\
MH04                  & 
\begin{tabular}[c]{@{}c@{}}\rednote{15.201$\pm$15.92}\\\rednote{ (10.145)}\end{tabular} & \begin{tabular}[c]{@{}c@{}}6.229$\pm$10.01\\ (1.260)\end{tabular} &
\begin{tabular}[c]{@{}c@{}}6.427$\pm$9.93\\ (1.468)\end{tabular} & \begin{tabular}[c]{@{}c@{}}\bluenote{2.994$\pm$4.88}\\\bluenote{(0.913)}\end{tabular} &  & \begin{tabular}[c]{@{}c@{}}\rednote{2.605$\pm$2.81}\\\rednote{(1.669)}\end{tabular} & \begin{tabular}[c]{@{}c@{}}2.301$\pm$3.64\\ (0.747)\end{tabular} & \begin{tabular}[c]{@{}c@{}}2.239$\pm$3.53\\ (0.735)\end{tabular} &
\begin{tabular}[c]{@{}c@{}}2.221$\pm$3.49\\ (0.735)\end{tabular}& \begin{tabular}[c]{@{}c@{}}\bluenote{1.260$\pm$1.34}\\\bluenote{(0.772)}\end{tabular} \\
MH05                        & 
\begin{tabular}[c]{@{}c@{}}\rednote{13.922$\pm$15.47}\\\rednote{(9.143)}\end{tabular} &
\begin{tabular}[c]{@{}c@{}}5.928$\pm$15.00\\ (1.374)\end{tabular} & \begin{tabular}[c]{@{}c@{}}4.847$\pm$9.92\\(1.497)\end{tabular} & \begin{tabular}[c]{@{}c@{}}\bluenote{1.753$\pm$2.47}\\\bluenote{(0.686)}\end{tabular} &  & \begin{tabular}[c]{@{}c@{}}\rednote{2.376$\pm$3.08}\\\rednote{ (1.090)}\end{tabular} & \begin{tabular}[c]{@{}c@{}}1.660$\pm$2.59\\ (0.598)\end{tabular} &
\begin{tabular}[c]{@{}c@{}}1.885$\pm$3.14\\ (0.595)\end{tabular} & \begin{tabular}[c]{@{}c@{}}1.951$\pm$3.34\\(0.597)\end{tabular} & \begin{tabular}[c]{@{}c@{}}\bluenote{0.969$\pm$1.27}\\\bluenote{(0.443)}\end{tabular} \\
V101                       & 
\begin{tabular}[c]{@{}c@{}}\rednote{25.256$\pm$27.57}\\ \rednote{(16.314)}\end{tabular} &
\begin{tabular}[c]{@{}c@{}}23.694$\pm$38.42\\ (8.886)\end{tabular} & \begin{tabular}[c]{@{}c@{}}24.674$\pm$32.08\\(10.596)\end{tabular} & \begin{tabular}[c]{@{}c@{}}\bluenote{9.682$\pm$13.43}\\\bluenote{(5.475)}\end{tabular} &  & \begin{tabular}[c]{@{}c@{}}17.598$\pm$20.90\\(9.802)\end{tabular} & \begin{tabular}[c]{@{}c@{}}13.402$\pm$19.10\\ (5.417)\end{tabular} & \begin{tabular}[c]{@{}c@{}}17.400$\pm$25.00\\ (6.230)\end{tabular} &
\begin{tabular}[c]{@{}c@{}}\rednote{17.844$\pm$25.86}\\\rednote{(6.230)}\end{tabular} & \begin{tabular}[c]{@{}c@{}}\bluenote{5.732$\pm$9.19}\\\bluenote{(3.008)}\end{tabular} \\
V102                        & 
\begin{tabular}[c]{@{}c@{}}\rednote{28.821$\pm$33.85}\\ \rednote{(15.540)}\end{tabular} &
\begin{tabular}[c]{@{}c@{}}24.904$\pm$45.06\\ (4.680)\end{tabular} &
\begin{tabular}[c]{@{}c@{}}26.179$\pm$39.96\\(7.292)\end{tabular} & \begin{tabular}[c]{@{}c@{}}\bluenote{6.548$\pm$9.65}\\\bluenote{(2.651)}\end{tabular} &  & \begin{tabular}[c]{@{}c@{}}22.823$\pm$38.48\\(9.039)\end{tabular} & \begin{tabular}[c]{@{}c@{}}\rednote{24.537$\pm$43.96}\\\rednote{ (5.196)}\end{tabular} & \begin{tabular}[c]{@{}c@{}}20.791$\pm$39.68\\ (5.332)\end{tabular} &
\begin{tabular}[c]{@{}c@{}}20.973$\pm$39.66\\ (5.428)\end{tabular} & \begin{tabular}[c]{@{}c@{}}\bluenote{4.841$\pm$5.74}\\\bluenote{(2.475)}\end{tabular} \\
V103                       & 
\begin{tabular}[c]{@{}c@{}}17.162$\pm$19.95\\ (9.932)\end{tabular} &
\begin{tabular}[c]{@{}c@{}}15.725$\pm$30.72\\ (3.755)\end{tabular} & \begin{tabular}[c]{@{}c@{}}\rednote{24.530$\pm$52.15}\\\rednote{(4.757)}\end{tabular} & \begin{tabular}[c]{@{}c@{}}\bluenote{6.410$\pm$9.15}\\\bluenote{(2.027)}\end{tabular} &  & \begin{tabular}[c]{@{}c@{}}\rednote{13.051$\pm$13.45}\\\rednote{(8.663)}\end{tabular} & \begin{tabular}[c]{@{}c@{}}10.602$\pm$14.66\\ (4.226)\end{tabular} & \begin{tabular}[c]{@{}c@{}}8.516$\pm$12.07\\ (3.584)\end{tabular} &
\begin{tabular}[c]{@{}c@{}}8.441$\pm$11.94\\ (3.560)\end{tabular} & \begin{tabular}[c]{@{}c@{}}\bluenote{6.457$\pm$9.22}\\\bluenote{(2.674)}\end{tabular} \\
V201                       & 
\begin{tabular}[c]{@{}c@{}}\rednote{5.090$\pm$4.36}\\ \rednote{(3.877)}\end{tabular} &
\begin{tabular}[c]{@{}c@{}}2.771$\pm$4.57\\ (1.110)\end{tabular} & \begin{tabular}[c]{@{}c@{}}2.617$\pm$3.53\\(1.217)\end{tabular} & \begin{tabular}[c]{@{}c@{}}\bluenote{1.488$\pm$1.72}\\\bluenote{(0.747)}\end{tabular} &  & \begin{tabular}[c]{@{}c@{}}3.784$\pm$4.22\\(2.375)\end{tabular} & \begin{tabular}[c]{@{}c@{}}4.486$\pm$7.77\\ (1.400)\end{tabular} & \begin{tabular}[c]{@{}c@{}}5.421$\pm$10.84\\ (1.401)\end{tabular} &
\begin{tabular}[c]{@{}c@{}}\rednote{5.664$\pm$11.64}\\\rednote{ (1.407)}\end{tabular} & \begin{tabular}[c]{@{}c@{}}\bluenote{1.473$\pm$1.63}\\\bluenote{(0.878)}\end{tabular} \\
V202                        & 
\begin{tabular}[c]{@{}c@{}}\rednote{11.916$\pm$11.99}\\\rednote{(8.187)}\end{tabular} &
\begin{tabular}[c]{@{}c@{}}11.250$\pm$22.75\\(2.764)\end{tabular} & \begin{tabular}[c]{@{}c@{}}10.486$\pm$20.86\\(2.841)\end{tabular} & \begin{tabular}[c]{@{}c@{}}\bluenote{4.308$\pm$5.47}\\\bluenote{(1.890)}\end{tabular} &  & \begin{tabular}[c]{@{}c@{}}8.694$\pm$9.44\\(5.637)\end{tabular} & \begin{tabular}[c]{@{}c@{}}7.329$\pm$10.264\\ (3.300)\end{tabular} & \begin{tabular}[c]{@{}c@{}}8.677$\pm$14.22\\ (3.062)\end{tabular} &
\begin{tabular}[c]{@{}c@{}}\rednote{8.760$\pm$14.42}\\\rednote{(3.071)}\end{tabular} & \begin{tabular}[c]{@{}c@{}}\bluenote{3.539$\pm$4.07}\\\bluenote{(1.883)}\end{tabular} \\
\hline \hline
\end{tabular}
}
\end{table}

\newpage

\section{Conclusion}
\label{conclusion}
In this paper, we have proposed a lightweight pose correction algorithm for relative frames between keyframes that can be easily integrated into existing keyframe-based SLAM systems. The algorithm was derived by preserving the measurement constraints of two updated keyframes and utilizing the notion that the measurement observed in both keyframes remains constant regardless of the update. By doing so, the algorithm avoids singularities and numerical sensitivity that existing interpolation-based methods suffer from. The algorithm was applied to poses generated from the current state-of-the-art ORB-SLAM2 in KITTI and EuRoC datasets. The algorithm demonstrated results superior to the existing interpolation methods in both translation and rotation for all three datasets. The computation time of the proposed algorithm was only a few milliseconds longer than the baseline methods, which is negligible in the overall SLAM process. Applications requiring visual information that may appear in non-keyframes can benefit from the proposed algorithm with negligible cost to computation time. Since the proposed module can be easily attached to existing keyframe-based SLAM systems, the algorithm may be used in a wide range of fields.

\bibliographystyle{bibtex/splncs}
\bibliography{bibtex/egbib}

\end{document}